\documentclass[conference]{IEEEtran}
\IEEEoverridecommandlockouts
\usepackage{cite}
\usepackage{amsmath,amssymb,amsfonts}
\usepackage{algorithm}
\usepackage{algpseudocode}
\usepackage{graphicx}
\usepackage{color}
\usepackage{textcomp}
\usepackage{multirow}
\usepackage{comment}
\usepackage{tikz}
\usepackage{textcomp}
\usepackage{hyperref}
\usepackage{lipsum}

\newcommand\copyrighttext{%
  \footnotesize \textcopyright 2021 IEEE.  Personal use of this material is permitted.  Permission from IEEE must be obtained for all other uses, in any current or future media, including reprinting/republishing this material for advertising or promotional purposes, creating new collective works, for resale or redistribution to servers or lists, or reuse of any copyrighted component of this work in other works.
  DOI: 10.1109/IIAI-AAI53430.2021.00010}
\newcommand\copyrightnotice{%
\begin{tikzpicture}[remember picture,overlay]
\node[anchor=south,yshift=10pt] at (current page.south) {\fbox{\parbox{\dimexpr\textwidth-\fboxsep-\fboxrule\relax}{\copyrighttext}}};
\end{tikzpicture}%
}

\def\BibTeX{{\rm B\kern-.05em{\sc i\kern-.025em b}\kern-.08em
    T\kern-.1667em\lower.7ex\hbox{E}\kern-.125emX}}
\begin{document}

\title{Development of an Extractive Title Generation\\System Using Titles of Papers of Top Conferences\\for Intermediate English Students}

\author{\IEEEauthorblockN{Kento Kaku, Masato Kikuchi, Tadachika Ozono, Toramatsu Shintani}
\IEEEauthorblockA{\textit{Department of Computer Science, Graduate School of Engineering} \\
\textit{Nagoya Institute of Technology}\\
Gokiso-cho, Showa-ku, Nagoya, Aichi, 466-8555, Japan\\
E-mail:\{kaku,kikuchi,ozono,tora\}@ozlab.org}}

\maketitle
\copyrightnotice{}

\begin{abstract}
    The formulation of good academic paper titles in English is challenging for intermediate English authors (particularly students). This is because such authors are not aware of the type of titles that are generally in use. We aim to realize a support system for formulating more effective English titles for intermediate English and beginner authors. This study develops an extractive title generation system that formulates titles from keywords extracted from an abstract. Moreover, we realize a title evaluation model that can evaluate the appropriateness of paper titles. We train the model with titles of top-conference papers by using BERT. This paper describes the training data, implementation, and experimental results. The results show that our evaluation model can identify top-conference titles more effectively than intermediate English and beginner students.
\end{abstract}


\section{Introduction}
The title of a paper is a concise and precise description of its contents. Therefore, it is important to determine an effective title for a paper. However, for those beginning to write papers, such as intermediate English and beginner students(hereinafter, referred to as just ``students''), it is difficult to develop a paper title. The purpose of this study is to support students to develop a paper title by providing them with appropriate paper titles. In this paper, we propose a system for generating paper titles by applying an extractive document summarization method to the abstracts of papers. First, we extract keywords that are likely to be included in the paper's title from the abstract. Then, we generate title parts, which are components of the title. We obtain a suitable paper title candidate by arranging these title parts, examining the grammatical syntax, and evaluating the paper title. We use {\em bidirectional encoder representations from transformers}\cite{Devlin} (BERT) to extract keywords and evaluate titles.

The remainder of this paper is organized as follows: Section 2 shows certain related works. Section 3 explains the extractive paper title generation method using BERT and the implementation of the system. Section 4 presents details of the experiments and a discussion of the system. Finally, Section 5 concludes the paper.

\section{Related Work}
\subsection{Traditional Title Generation Methods}
Statistical machine translation methods have been conventionally used for generating titles\cite{Banko}\cite{Jin}. To generate titles, it is necessary to aggregate important information scattered in multiple sentences into a shorter sentence. Banko et al. conducted research on title generation\cite{Banko}. They focused on the fact that headline-style short summaries cannot be achieved by sentence-level extractive summarization methods. A statistical model was used to select the words to be included in the title and to order the selected words. The word sequence with the highest scores for both word selection and word order was adopted as the title.

However, this framework has two problems. One is the tendency to prefer word sequences containing common words during the word ordering phase. The other is that the consideration of content and common words as being equivalent during the word selection phase may result in a decrease in the quality of the generated titles. Jin et al. attempted to solve these problems by introducing a hidden state before selecting the words to be included in the title from a document\cite{Jin}.

\subsection{Existing Methods Using DNN}
There are two types of methods for document summarization: extractive type and abstractive type. Extractive summarization extracts words and sentences that are considered particularly important from the original document and arrange these to obtain a summary text. Abstractive summarization generates summary sentences by comprehending the meaning of the source document and generating the summary similarly as a human, using words in the dictionary. Research on extractive summarization is highly active. Research on abstractive summarization has also been active because neural network-based {\em encoder-decoder models} have become available and sentence generation has become feasible. In our study, we use extractive summarization.

Rush et al. proposed a method using the {\em attention-based summarization} (ABS) model for abstractive summarization by incorporating an {\em attention mechanism}\cite{Bahdanau} into the {\em encoder--decoder model}\cite{Rush}. As an extractive summarization using deep learning, Cheng et al. proposed a method to generate summary sentences. Herein, the sentences to be extracted are determined by assigning scores to the sentences in the original document using RNN\cite{Cheng}. 

Conventionally, statistical methods have been used for grammatical error detection and correction. Rei et al. proposed a neural network-based method for grammatical error detection\cite{Rei}. In this method, grammatical error detection is achieved by training a {\em bidirectional LSTM} with error labels attached to words in the grammatical error portion of the learner corpus. In addition, Yuan et al. proposed a method for correcting grammatical errors using neural machine translation\cite{Yuan}. This method enables grammatical error correction by training the {\em encoder--decoder model} using RNN and {\em attention} with the sentences before and after grammatical error correction in the learner corpus. Because there is no learner corpus of paper titles, in this study, we trained BERT to detect grammatical errors by learning grammatically incorrect sentences from actual paper titles with different word orders.

\subsection{Title Generation Method Using DNN}
After describing the difference between ordinary document summarization and paper title generation, Ohbe et al. proposed an abstractive and extractive paper title generation method that uses RNN to generate paper titles from abstracts\cite{Ohbe}. In this method, the paper title is generated from the abstract. This is similar to the approach adopted in the present study. According to Ohbe et al., document summarization requires the simultaneous summarization of a large number of documents as well as the readability of the summarized text. Meanwhile, generated paper titles need not necessarily be grammatically correct. This is because it does not require the simultaneous processing of a large number of papers and because manual correction is considered. In addition, templates are used considering the availability of standardized phrases for paper titles.

Mishra et al.\cite{Mishra} proposed a paper title generation method using GPT-2, which is a {\em transformer-based natural language processing model} similar to BERT. Because GPT-2 has a probabilistic characteristic, its output varies each time. Multiple title candidates are generated using this property, and the best one is selected from among these. The ``relevancy scores'' of the improved title candidate and the unimproved one are compared, and the one with the higher value is used as the final output title. In this study, we also generate multiple title candidates and select the more appropriate title candidates from among these.

The position of this research is described below. Ohbe et al.\cite{Ohbe} proposed extractive and abstractive methods for generating paper titles. However, both these can be improved. The abstractive method uses the dictionary of a natural language processing model and outputs sentences that appear to be titles but are nevertheless undesirable. This is because it tends to output words that are not related to the input abstract. In the experiment, the extracted method yield better results. Therefore, we consider the extractive method as being more potential than the abstractive method. Hence, in this study, we investigate an extractive paper title generation method. Furthermore, we focus on the method of creating a paper title from extracted keywords because it is important for the extractive paper title generation method.

\section{Paper Title Generation System}
This chapter explains how our system generates a title from an abstract by the following three steps. First, our system extracts keywords from a given abstract. Next, the system arranges the keywords as title candidates. Finally, it evaluates the candidates to select the best title. 

There are two problems that need to be solved to achieve this. (1) Because a paper title consists of approximately 5--20 words, the maximum number of words of title candidates is approximately 20. Therefore, the evaluation of all title candidates would consume a long time. (2) Grammatical syntax is not considered while evaluating titles. Therefore, grammatically incorrect title candidates are also evaluated. To solve Problem (1), we introduce title parts. A title part is a sequence of words that are likely to be included in the title of a paper. The number of elements to be reordered and the number of generated title candidates are reduced by reordering the title parts to generate title candidates. This, in turn, reduces the execution time of title evaluation. In addition, it is possible to address technical terms composed of general words by using title parts. To solve Problem (2), we introduce a method called grammar checking. We use grammatical patterns for grammar checking. A grammatical pattern is a part of the syntax tree of a paper title that consists only of phrases and clauses, except the tags representing parts of speech of words (leaves of the syntax tree). The grammatical patterns of actual paper titles are collected in advance. Then, these are matched with the grammatical patterns of title candidates during title generation. The title candidate is considered to be grammatically correct if any of the grammatical patterns matches the grammatical pattern of the title candidate, and to be grammatically incorrect if there is no match. This enables us to verify the grammatical correctness. In addition, the grammatical check enables us to eliminate unnecessary title candidates before title evaluation. This reduces the execution time of title evaluation.

\subsection{Architecture}
\begin{figure}[t]
    \centering
    \includegraphics[width=8.8cm]{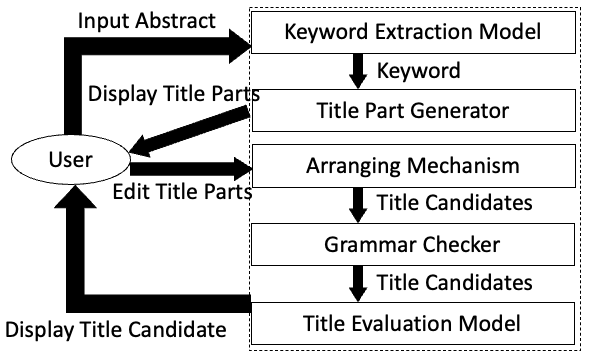}
    \caption{System architecture}
    \label{system_configuration}
\end{figure}

The system architecture is shown in Fig. \ref{system_configuration}. The system consists of a keyword extraction module, title candidate generation module, and title evaluation module. First, the user inputs into the client the abstract of the paper for which the title is to be generated. The abstract is sent to the server, where it is input into the keyword extraction model. The keywords extracted from the abstract are input to the title part generator. The title parts generated are sent to the client. The user can examine and modify these if necessary. The modified title parts are sent to the server and input to the arranging mechanism. The arranging mechanism generates title candidates by arranging the title parts and inputs the generated title candidates to the grammar checker. The grammar checker eliminates the grammatically incorrect title candidates and outputs only the grammatically correct ones. These title candidates are input to the title evaluation model. The model evaluates the validity of the title and outputs only those whose evaluation value is higher than a threshold. The output title candidates are sent to the client and presented to the user. The following describes the keyword extraction module, title candidate generation module, and title evaluation module.

\subsection{Keyword Extraction Based on BERT}
The keyword extraction module consists of a keyword extraction model. The keyword extraction module extracts keywords that are likely to be included in the paper title from the abstract.

\begin{figure}[t]
    \centering
    \includegraphics[width=8.8cm]{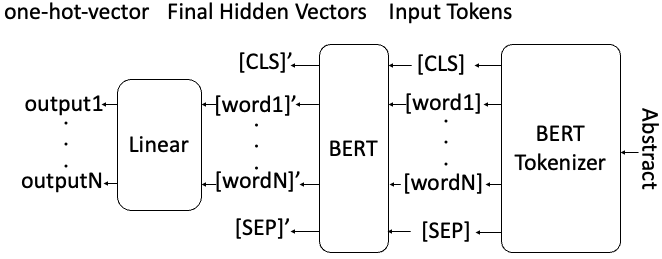}
    \caption{Keyword extraction model}
    \label{extraction_model}
\end{figure}

The keyword extraction model extracts words that are likely to be included in the title of the paper by binary classification based on whether each word in the abstract appears in the title of the paper. The keyword extraction model is developed using BERT. Fig. \ref{extraction_model} illustrates the keyword extraction model. First, for BERT to be capable of handling abstracts, the abstract strings are converted into an input sequence of word IDs using BERT tokenizer. BERT tokenizer divides words into subwords and converts these into word IDs. It then inserts a [CLS] token at the beginning of the input sequence and a [SEP] token at the end of each sentence. These are special tokens that represent the meaning of the input sequence and the sentence break, respectively. We now enter the input sequence of the abstract into BERT and obtain its output. A vector of scores for each word in the abstract is generated by performing a linear transformation on the vector of hidden representations of each word in the abstract except for the special tokens. The system classifies the words into those that appear in the title of the paper and those that do not by comparing these scores with a threshold value. During the training, tokens in the abstract that were included in the title were assigned one (the label for correct answer), and those not included in the title were assigned zero. In this study, we trained a keyword extraction model using 9,229 paper titles and abstracts of AAAI (a top conference) that were collected using the DBLP API.

This result reveals that most of the words in titles are included in abstracts. Specifically, the abstracts of the 9,732 AAAI papers we used as data contained 85\% words in their titles. Therefore, the method of title generation using words in the abstract is considered to be effective.

\subsection{Title Candidate Generation}
The title candidate generation module consists of a title part generator and an arranging mechanism. In the title candidate generation module, title candidates are generated from keywords.

The title part generator generates title parts based on the abstract and keywords. In this study, the title part is the word sequence of the part of the abstract where the keywords are consecutive. The title part generator is explained using Algorithm \ref{title_parts_generator}. First, the word sequence with the longest matching keyword is extracted from the abstract. At this time, the words in the abstract are split into subwords by the BERT tokenizer. A word may be broken down into multiple subwords. Such subwords have a marker ``\#\#'' at the beginning of the token, except at the beginning of the word. If the marker is present, it is removed, and the token is merged with the previous subword. If the marker is not present, a space is provided between the token and previous subword. The unnecessary title parts are excluded from the title parts generated by this process. Unnecessary title parts are those that match with other title parts and those that consist of only ``-.'' Finally, the user can review the title parts and edit these directly if necessary. The title parts are generated by the title part generator by these processes.

\begin{algorithm}[t]
    \caption{Pseudo-code of title part generator}
    \label{title_parts_generator}
    \begin{algorithmic}[1]
        \Function{GenerateTitleParts}{keywords, abstract}
            \State title\_parts = \Call{GetLM}{keywords, abstract} \\
            \Comment{Extract the longest matching word sequence.}
            \State title\_parts = \Call{Repair}{title\_parts} \\
            \Comment{Join the split words again.}
            \State title\_parts = \Call{Dump}{title\_parts} \\
            \Comment{Exclude unnecessary title parts.}
            \State title\_parts = \Call{UserEdit}{title\_parts} \\
            \Comment{Perform user editing.}
            \State \Return title\_parts
        \EndFunction
    \end{algorithmic}
\end{algorithm}

The arranging mechanism generates paper title candidates by arranging title parts. We explain the reordering mechanism using Algorithm \ref{title_shape_algorithm}. In the initial state, the arguments are a set of title parts, an empty string that would subsequently become a title candidate, and an empty array that stores a list of title candidates. First, a title part is extracted from the set of title parts and combined with the string of title candidates. This process is repeated by recursive calls of the shape function until the set of title parts is empty. Then, the title candidate is considered to be complete, and the grammar check and title evaluation are performed to determine its suitability as a title. The title candidates that are assessed to be suitable as titles are added to the array that stores the title candidates. Appropriate title candidates can be obtained by performing these processes for all the permutations of the title parts.

\begin{algorithm}[t]
    \caption{Pseudo-code for arranging mechanism}
    \label{title_shape_algorithm}
    \begin{algorithmic}[1]
        \Function{Shape}{parts, title, title\_list}
            \For{$i=1$ to $\Call{Length}{parts}$}
                \State \Call{Add}{title, parts[i]}
                \State \Call{Delete}{parts, parts[i]}
                \If{$\Call{Length}{parts} == 0$} 
                    \If{\Call{GrammarCheck}{title}} \\
                    \Comment{Determine whether it is grammatically correct.}
                        \State score = \Call{Evaluate}{title} \\
                        \Comment{Determine whether the title is appropriate.}
                        \If{$score \geq 0.5$} \\
                        \Comment{Add title candidates to title\_list}
                            \State \Call{Add}{title\_list, score, title}
                        \EndIf
                    \EndIf
                \Else \Comment{Call this recursively if parts remain}
                    \State title\_list = \Call{Shape}{parts, title, title\_list}
                \EndIf
            \EndFor
            \State \Return title\_list
        \EndFunction
    \end{algorithmic}
\end{algorithm}

\subsection{Title Evaluation Based on BERT}
The title evaluation module consists of a grammar checker and title evaluation model. The title evaluation module evaluates the title candidates and selects the one with the highest evaluation value from among these.

The grammar checker examines the grammatical syntax of the title candidate based on the structure of the syntax tree of the paper title. Here, we use a grammar pattern. A grammar pattern is a part of the syntax tree that consists only of tags representing phrases and clauses, excluding the leaves of the tree such as words and POS tags. The grammatical patterns are collected in advance from actual paper titles, and the grammatical syntax of a title candidate is determined by matching the grammatical patterns of the actual paper title with those of the title candidate. If at least one of the collected grammatical patterns matches with those of the title candidate, the title candidate is considered to be grammatically correct and entered into the title evaluation model. If none of the collected paper titles match the grammatical pattern of the title candidate, the title candidate is considered to be grammatically incorrect and excluded. A syntactic analyzer is used to generate a syntactic tree from the actual paper title and title candidates. In this study, we use Berkeley Parser.

\begin{figure}[t]
    \centering
    \includegraphics[width=8.8cm]{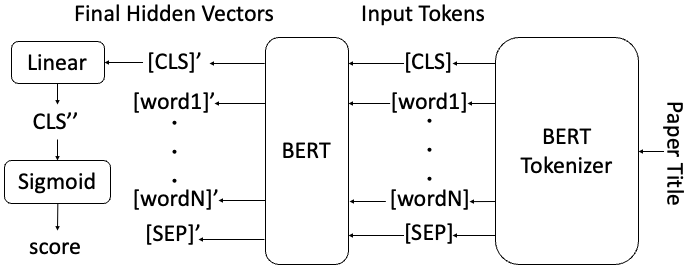}
    \caption{Title evaluation model}
    \label{evaluation_model}
\end{figure}

The title evaluation model evaluates the appropriateness of a candidate title as a title of a paper by performing binary classification and comparing the evaluated value with a threshold value. This enables us to obtain only paper title candidates to be presented to the user. The title evaluation model is developed using BERT. Fig. \ref{evaluation_model} illustrates the title evaluation model. First, we convert the paper title candidates into an input sequence using the BERT tokenizer as in the keyword extraction model. Next, the input sequence of the paper title candidates is input to BERT, and the output is obtained. The output is a vector of latent representations for each token. However, we assume that among these latent representations, the [CLS] token that captures the entire meaning of the sentence has information on the validity of the paper title. Therefore, we linearly transform the latent expression vectors of the [CLS] tokens to obtain the scores of the paper title candidates. These scores are input into the sigmoid function to be converted into scores with values in the range zero--one. By comparing these values with a threshold value, we classify the candidates into those that are appropriate as paper titles and those that are not. We assign one as the correct answer label for the values before inputting these into the sigmoid function for actual paper titles, and zero for sentences that are not paper titles. For training the title evaluation model, we use the titles of 9,232 papers from AAAI (a top conference) collected using the DBLP API. Actual paper titles are used as positive examples for training. In addition, the first sentence of the abstract (an example of a normal sentence) and sentences generated by randomly rearranging the word order of actual paper titles (examples of grammatically incorrect paper titles) are trained as negative examples.

An example of title generation by this system is shown in Table \ref{example}. The table shows that user's modification of the title parts can promote incomplete keyword extraction. This causes the words in the title parts to be identical to those in the actual title. However, the actual title does not rank well. This may be because the title evaluation model does not consider the meaning of the abstract. It is necessary to develop a title evaluation model that does so.

\begin{table*}[t]
    \centering
    \caption{Example of titles generated by the system(C. Stachniss, W. Burgard, AAAI, 2005)}
    \label{example}
    \begin{tabular}{|l|} \hline
        {\bf Actual title:} Mobile Robot Mapping and Localization in Non-Static Environments\\
        {\bf Title parts generated by system:} ``- static,'' ``of mapping,'' ``dynamic,'' ``a mobile robot,'' ``mobile robot in,'' ``for mapping and local''\\
        {\bf Title parts edited by user:} ``mobile robot,'' ``in,'' ``mapping and localization,'' ``non - static,'' ``environments''\\
        {\bf Title candidate 1:} mapping and localization in non-static mobile robot environments / score: 0.849137544631958\\
        {\bf Title candidate 2:} mobile robot environments in non-static mapping and localization / score: 0.7966495752334595\\
        {\bf Title candidate 3:} mapping and localization environments in non-static mobile robot / score: 0.7281235456466675\\
        \hspace{10mm}\vdots \\
        {\bf Title candidate 7:} mobile robot mapping and localization in non-static environments / score: 0.6613911390304565\\
        \hspace{10mm}\vdots \\ \hline
    \end{tabular}
\end{table*}

\section{Experiments}
\subsection{Experiment on Title Generation System}
We conducted an evaluation experiment to compare our system with existing summarization methods by ROUGE-1, –L and BLEU-2. In addition, we calculated the number of words in the titles generated by each method. We used 373 AAAI paper titles and abstracts in this experiment. Because the number of words extracted by our keyword extraction model varies substantially depending on the input, we set the fundamental threshold as -0.5 and adjusted the threshold so that the number of title parts generated was between three and six. The title parts were not edited by humans, and the generated title candidate with the highest score in the title evaluation model was adopted. When all the title candidates were rejected by the grammar checker, the title parts in their original order were adopted. We also compared our system with LexRank\cite{Erkan} and TextRank\cite{Mihalcea}. Both the systems were implemented using the Python package Sumy\footnote{Sumy - https://github.com/miso-belica/sumy}. The number of summary sentences to be generated was set to one.

\begin{table}
    \centering
    \caption{ROUGE and BLEU scores of each method}
    \label{ROUGE_BLEU}
    \begin{tabular}{|c|c|c|c|c|} \hline 
        \multirow{4}{*}{Method} & \multirow{2}{*}{\begin{tabular}{c}the average of\\the number of words\\in the generated\\summaries or titles\end{tabular}} &\multicolumn{2}{|c|}{}& \\ 
        &&\multicolumn{2}{|c|}{ROUGE}&BLEU \\ \cline{3-5}
        &&&& \\
        && 1 & L & 2 \\ \hline
        TextRank & 33.1 & 0.0456 & 0.0477 & 0.1122 \\ 
        LexRank & 23.7 & 0.0613 & 0.0597 & 0.1301 \\
        Our System & 7.9 & 0.0558 & 0.0566 & 0.1197 \\ \hline
    \end{tabular}
\end{table}

The experimental results are shown in Table \ref{ROUGE_BLEU}. The results of ROUGE-1, -L and BLEU-2 in the table show that our system is better than TextRank, but worse than LexRank. However, our system is not significantly different from the other methods. The summaries generated by the other methods are too long for titles. Our method outperforms the others in terms of title generation because it achieved similar scores with fewer words. Moreover, our method is able to generate more title-like sentences.

\subsection{Experiment on Title Evaluation Model}
We trained a title evaluation model using 9,232 paper titles from AAAI collected using the DBLP API. The actual title of the paper was used as a positive example. The first sentence of the abstract and a randomly altered word order of the actual title were used as negative examples of a normal sentence and a grammatically incorrect title, respectively. We conducted an experiment to evaluate the title evaluation model using 30 AAAI paper titles, 778 IJCAI paper titles, 30 JSAI paper titles, and 30 randomly arranged word orders of AAAI paper titles. The AAAI and IJCAI paper titles were collected using the DBLP API, and the JSAI paper titles were collected using CiNii. The IJCAI paper titles were used because we used AAAI paper titles for training and therefore, had to compare these with the results of the evaluation of paper titles from other top conferences. The JSAI paper titles were used to compare the results of the evaluation of Japanese conference paper titles with those of the top conferences. We also asked three second-year graduate students and four fourth-year undergraduate students (all of them were Japanese intermediate English students) to evaluate the AAAI paper titles and the titles in which the word order was altered. This enabled us to compare the evaluation results of test users with those of the title evaluation model. In the data used in this experiment, the system names were deleted if these were separated by ``:'' at the beginning of the paper title. TABLE \ref{evaluation_evaluation} shows the percentages of the titles that were assessed to be appropriate as paper titles. The ratio is defined as follows:
\[ratio = \frac{number\ of\ paper\ titles\ deemed\ appropriate}{number\ of\ paper\ titles\ used\ in\ the\ experiment}\]

\begin{table}[t]
    \centering
    \caption{Experimental results of the title evaluation model}
    \label{evaluation_evaluation}
    \begin{tabular}{|c|c|c|c|} \hline
      & Title evaluation model & M2 & B4 \\ \hline
      AAAI(top conference) & 0.93 & 0.73 & 0.63 \\ \hline
      IJCAI(top conference) & 0.88 & - & - \\ \hline
      JSAI(not top conference) & 0.70 & - & - \\ \hline
      \begin{tabular}{c}
        AAAI titles with\\different word order
      \end{tabular}
       & 0.47 & 0.16 & 0.36 \\ \hline
    \end{tabular}
\end{table}

Table \ref{evaluation_evaluation} shows that the title evaluation model assigns higher ratings to the titles of the top conference papers than test users do. In addition, because AAAI paper titles were used for training, it is likely that the title evaluation model rated these higher than IJCAI paper titles. However, there was no significant difference between the two results. This indicates that the title evaluation model rated the top-conference paper titles higher. In addition, the title evaluation model had a marginally lower result of 0.70 in the assessment of the titles of JSAI papers as appropriate. This indicates that the title evaluation model learned the titles of AAAI papers and assigned a higher score to the more native paper title. However, the percentage of assessments wherein the title evaluation model, second-year graduate students, and fourth-year graduate students determined the titles of the papers with different word orders as appropriate was 0.47, 0.16, and 0.36, respectively. This indicates that the title evaluation model assigns higher ratings to grammatically incorrect paper titles than test users do. These results indicate that the title evaluation model is effective for evaluating grammatically correct paper titles. Furthermore, this system may be effective for intermediate English students. These results also indicate that when our system is trained on papers in a particular field, our system may be capable of aiding researchers unfamiliar with that field to formulate titles for papers in that field. However, a problem is that the system cannot evaluate grammatical errors in paper titles. This experiment was performed with a marginal amount of data because it was a preliminary experiment. We plan to perform another experiment with a large amount of data.

\subsection{Experiment of Grammar Check Module}
Three AAAI paper titles randomly selected from the collected papers were divided into five title parts and arranged to generate a total of 360 paper title candidates. The grammatical syntax of these candidate titles was examined by the grammar checker, and the results were manually compared to evaluate the grammar checker. We collected grammatical patterns from 9,232 AAAI paper titles in advance, and the grammar checker verified the correctness of the grammatical patterns by matching these with the grammatical patterns of the paper title candidates. The results of this experiment are shown in Table \ref{grammer_check_evaluation}. TP, FP, FN, and TN in Table \ref{grammer_check_evaluation} represent true positive, false positive, false negative, and true negative, respectively.

\begin{table}[t]
    \centering
    \caption{Experimental results of grammar checker. Note that almost all the test sentences, which are permutations of extracted keywords, are grammatically incorrect.}
    \label{grammer_check_evaluation}
    \begin{tabular}{|@{\hspace{0mm}}c@{\hspace{0mm}}|c|c|c|}\hline
        && \multicolumn{2}{|c|}{Manual judgment} \\ \hline
        && Positive & Negative \\ \hline
        \multirow{2}{*}{\begin{tabular}{c}Assessment by\\grammar checker\end{tabular}}&Positive&\begin{tabular}{c}TP\\0.14\end{tabular}&\begin{tabular}{c}FP\\0.36\end{tabular}\\ \cline{2-4}
        &Negative&\begin{tabular}{c}FN\\0.03\end{tabular}&\begin{tabular}{c}TN\\0.47\end{tabular}\\ \hline
    \end{tabular}
\end{table}

From Table \ref{grammer_check_evaluation}, we succeeded in reducing the number of grammatically incorrect paper title candidates by more than half without eliminating most of the grammatically correct ones. This indicates that the execution time of the system could be reduced significantly without altering the quality of the paper title candidates. However, we could not eliminate over 30\% of the grammatically incorrect paper title candidates. It is likely that the title evaluation model excluded ungrammatical paper title candidates to a certain extent, or that users did not adopt these as appropriate paper titles. However, there is scope for improvement in terms of a reduction in the execution time of title evaluation.

\section{Conclusions}
In this paper, we describe a prototype of an extractive paper title generation system using BERT to assist people who are not familiar with paper writing and paper title creation. Our experimental results showed that the title evaluation model was capable of evaluating paper titles more effectively than Japanese graduate students. Furthermore, the grammar checker was capable of eliminating unnecessary paper title candidates without altering the quality of the paper titles.

\section*{Acknowledgment}
This work was supported in part by JSPS KAKENHI Grant Numbers 19K12097, 19K12266.

\end{document}